\documentclass[11pt,a4paper]{article}
\usepackage[hyperref]{acl2019}
\usepackage{times}
\usepackage{tabularx}
\usepackage{latexsym}
\usepackage{adjustbox}
\usepackage{multirow}
\usepackage{url}
\usepackage{xcolor}
\usepackage{hyperref}
\usepackage{lipsum}
\usepackage{enumitem}
\usepackage{dirtytalk}

\usepackage{array,ragged2e}
\newcolumntype{P}[1]{>{\RaggedRight\arraybackslash}p{#1}}
\usepackage{caption}
\usepackage{graphicx}

\newcommand\blfootnote[1]{%
  \begingroup
  \renewcommand\thefootnote{}\footnote{#1}%
  \addtocounter{footnote}{-1}%
  \endgroup
}
\usepackage{colortbl}
\definecolor{mygray}{gray}{.9}
\aclfinalcopy 

\date{}

\begin{document}
\title{ErAConD: Error Annotated Conversational Dialog Dataset for Grammatical Error Correction}

\author{Xun Yuan$^{*2}$, Derek Pham$^{*1}$,  Sam Davidson$^{3}$ and Zhou Yu$^1$ \\$^1$Columbia University, $^2$Zhejiang University, $^3$University of California, Davis
\\ \texttt{\{dp3081, zy2461\}@columbia.edu} \\
\texttt{yuanxun@zju.edu.cn, ssdavidson@ucdavis.edu}}

\maketitle

\begin{abstract}
\blfootnote{$^{*}$Authors contributed equally to this work.}
Currently available grammatical error correction (GEC) datasets are compiled using essays or other long-form text written by language learners, limiting the applicability of these datasets to other domains such as informal writing and conversational dialog. In this paper, we present a novel GEC dataset consisting of parallel original and corrected utterances drawn from open-domain chatbot conversations; this dataset is, to our knowledge, the first GEC dataset targeted to a human-machine conversational setting.
We also present a detailed annotation scheme which ranks errors by perceived impact on comprehension, making our dataset more representative of real-world language learning applications.
To demonstrate the utility of the dataset, we use our annotated data to fine-tune a state-of-the-art GEC model. Experimental results show the effectiveness of our data in improving GEC model performance in a conversational scenario.

\end{abstract}

\section{Introduction}

In recent years, both researchers and businesses have attempted to build effective educational chatbots to help language learners improve their conversational skills in a second language (primarily English) \citep{huang2021chatbots}.
However, many such systems, such as GenieTutor Plus \citep{huang2017genietutorplus}, use rule-based dialog engines, and thus do not take advantage of recent developments in dialog generation using Transformer models, which have vastly improved the quality of modern chatbots \citep{liang2020gunrock}. Extant dialog systems for conversational language learning can be broadly classified into two types. In the first type, the chatbot serves as a teacher and repeatedly asks the user questions to test acquisition of specific words, syntax, and other pedagogical targets. In the second type, the chatbot serves as a conversational partner, encouraging users to chat with it and, in some cases, providing corrective feedback to learners \citep{fryer2020bots}. It is this latter type we hope to improve using our proposed dataset.

Grammatical error correction (GEC) models are needed to generate appropriate corrective feedback for this second type of educational chatbot. However, nearly all current GEC datasets focus on written essays, a domain which differs markedly from conversational speech in both syntax and style.
As a result, datasets drawn from written sources, such as student essays, produce poor results when applied to dialog \citep{davidson2019dependency}. There currently exists one dataset of error-annotated conversational utterances by English second language learners on which researchers can train and evaluate conversational GEC models, the Teacher-Student Chatroom Corpus \citep{caines-etal-2020-teacher}; this data is generated in the context of human-human interaction, specifically interactions between a teacher and a second language learner student. However, no similar conversational dataset focuses on the human-machine conversational setting. In this work we seek to address this gap in the available data by developing a high-quality, error-annotated dataset of learner dialog collected from an online educational chatbot.\ifaclfinal\footnote{Data is available at \url{https://github.com/yuanxun-yx/eracond}}\else\footnote{Data will be available on GitHub upon acceptance.}\fi
\ 
To appropriately annotate our data for language learning applications, we introduce a 3-level grammatical error classification structure in order to categorize errors based on severity. Our motivation for this error classification structure is to give users the opportunity to first focus on improving their most serious grammatical errors. To demonstrate the utility of the proposed dataset, we fine-tune and evaluate a state-of-the-art GEC model using our newly developed dataset.

\section{Related Work}

As with many NLP tasks, the current state-of-the-art in grammatical error correction involves using large Transformer-based language models such as BERT \citep{devlin2019bert}, RoBERTa \citep{liu2019roberta}, and XLNet \citep{yang2019xlnet}. To evaluate the utility of our dataset, we use \citet{omelianchuk2020gector}'s GECToR model, which reframes GEC as a sequence labelling task rather than a monolingual machine translation task. GECToR achieves SoTA results on the test corpus used for the BEA 2019 Shared Task on Grammatical Error Correction \citep{bryant2019bea}. 
Other promising supervised GEC models include those of \citet{stahlberg2021synthetic} and \citet{rothe2021simple}, who achieve SoTA results on the JFLEG \citep{napoles2017jfleg} and CoNLL-2014 \citep{ng2014conll} GEC datasets, respectively. Both models combine innovative synthetic data generation methods with large pretrained transformer language models.



Recent work related to the development of datasets for grammatical error correction include \citet{napoles2019enabling} who presents a dataset of native and non-native English writing. \citet{trinh2021new} proposes a new parallel dataset of Russian student writing. These datasets add to the growing number of GEC datasets available to the research community. However, as previously mentioned, no GEC dataset that contains chatbot-based human-machine conversational data, in English or any other language, is currently available. We seek to begin closing this gap with the present research.

\section{Data Collection}

\subsection{Data Collection Process}

We collected 186 dialogs containing 1735 user utterance turns of open-domain dialog data by deploying BlenderBot \citep{roller2020recipes} on Amazon Mechanical Turk (AMT) via LEGOEval. \citep{li_arnold_yan_shi_yu_2021}. We decided to deploy BlenderBot, because it is open-sourced and because it has relatively good coherence, and is known for its engagement and human-like conversational qualities. 

The AMT crowd-workers who conversed with our bot are L2 English speakers of at least intermediate proficiency.
The workers were asked to converse with our chatbot for at least 10 turns (a turn is defined as a bot/user utterance pair) either about movies or the COVID-19 pandemic; we chose these because of their universal experience and subjectivity of the two topics, resulting in a rich and diverse set of utterances in the dataset. Workers interacted with the bot using a typed interface, similar to a messaging app. We plan to expand this to an ASR-driven system as well as to additional conversational topics in future work.

\subsection{Data Annotation}
After collecting open-domain dialog data, we manually revised each user utterance to correct any non-standard or ungrammatical English usage. A subset of the dialogs are corrected by two annotators to provide multiple corrected targets for system evaluation--the remaining dialogs are corrected by a single annotator. Both annotators are graduate student native speakers of English.

\subsubsection{Annotation scheme}

We followed an annotation method similar to that proposed in \citet{naplava2022czech}, in which we asked annotators to revise any sentences containing ungrammatical elements. Our goal was to apply the minimum number of edits needed to make the utterance conform to standard written English while remaining as faithful to the source as possible. With this goal in mind, we designed our annotation scheme to conform to the rules of standard written English with two exceptions: internet shorthand and slang, and short responses which are incomplete sentences; both forms, while not acceptable in formal written English, are frequently considered acceptable in the context of informal dialog. 
We also made fluency edits \citep{napoles2017jfleg} of semantic and sentence construction errors, particularly those related to lexical choice, omission, and word order. For example, the source line \textit{``The movie tell about a poor girl that meet a prince and in love for him"}, suffers from non-native-like word choice. We corrected this utterance to \textit{``the movie tells about a poor girl that meets a prince and falls in love with him"}. We made these corrections with the intention of creating ground truth utterances which are as semantically and syntactically similar to the source as possible.

\subsubsection{Inter-annotator agreement}
We took several steps to ensure that annotators were meeting the goal of revising ungrammatical and disfluent sentences while retaining the semantic content of the source dialog. During our annotation training phase, we had both annotators correct an identical set of 26 dialogs. The annotators then reviewed each other's annotations on this subset and noted specific area of disagreement, which were then discussed with the lead researchers who provided specific instructions on how to resolve these discrepancies. Annotators then repeated the annotations of the same 26 dialogs to ensure that the provided instructions were being followed. During this process, we also asked annotators to note any changes which they believed changed the underlying semantics of a given dialog, so that such changes could be eliminated in subsequent passes. At each stage, we calculated inter-annotator agreement (as described below), and only continued with the annotation of the remaining 160 dialogs once our second-pass agreement levels were on-par with previously reported GEC corpora such as \citet{trinh2021new}.

For many ungrammatical sentences, there are multiple acceptable ways to correct the error. As a result GEC annotations can be quite variable, and traditional methods of calculating inter-annotator agreement are not informative \citep{rozovskaya2019grammar}. We therefore utilized two metrics for calculating inter-annotator agreement. The first, originally proposed by \citet{rozovskaya2010annotating}, asks each annotator to review and correct the corrections of the other annotators, and then calculates the percent of sentences which are unchanged on this second pass; these figures are shown as \say{Judged correct} in Table \ref{tb:annagree}. The second method, used in work such as \citet{trinh2021new} calculates the F${}_{0.5}$ by setting one annotator as reference and the other as hypothesis; these figures are also provided in Table \ref{tb:annagree}.

\begin{table}[htb]
\scriptsize
\centering
\begin{tabular}{ c|c|cccccc}
\hline 
\textbf{Ref} & 
\begin{tabular}{@{}c@{}}
\textbf{Judged}\\\textbf{Correct}
\end{tabular} &
\textbf{TP}& \textbf{FP} & \textbf{FN} & \textbf{Prec} & \textbf{Rec} & \textbf{F${}_{0.5}$} \\
\hline
0 & 95.9\% & 59 & 109 & 57 & 0.351 & 0.509 & 0.374 \\
1 & 96.2\% & 59 & 57 & 109 & 0.509 & 0.351 & 0.467 \\
\hline
\end{tabular}
\caption{\label{tb:annagree} 
Annotator agreement by F$_{0.5}$ score. Only dialogs with two annotators are compared. The first column indicates which annotator is selected as reference. The \say{Judged Correct} column indicates second-pass agreement between annotators.
}
\end{table}

\subsection{Error Types}
\label{error_types}

\begin{table}[htb]
\small
\centering
\begin{tabularx}{\linewidth}{>{\centering\hsize=.3\hsize\linewidth=\hsize}X>{\centering\hsize=.7\hsize\linewidth=\hsize}X>{\hsize=2\hsize\linewidth=\hsize}X} 
 \hline
 \textbf{Level} & \textbf{Impact on Meaning} & \textbf{Error Types} \\
 \hline
 1 & Trivial & Punctuation (excl. apostrophe) \& Casing \\ 
 \hline
 2 & Moderate & Acronyms, Abbreviations, Non-English Internet Slang, \& Apostrophe \\
 \hline
 3 & Significant & SV Agreement, Verb Form, Word Confusion, etc. \\ 
 \hline
\end{tabularx}
\caption{\label{tb:cat_errors}
Categorization of grammatical errors.}
\end{table}

\begin{table*}[htb]
\scriptsize
\centering
\begin{tabular}{ c| m{10cm} |l}
\hline
\textbf{Example} & \textbf{Message} & \textbf{Error} \\ \hline \hline
\rowcolor{mygray}
1 & yes, johnny depp, and brad pitt       & Punctuation \& Casing  \\ \hline
2 & Ok, what are you talking about? Kkkkkk       & Non-English Internet Slang  \\ \hline \rowcolor{mygray}
3 & I also like SF movies. It makes me think differently.       & Acronym  \\ \hline
4 & What's your fav movie right now?       &   Abbreviation \\ \hline \rowcolor{mygray}
5 & IT SEEMS DRAMATIC. ILL WATCH.       & Apostrophe   \\ \hline 
6 & She is not on the line now. Maybe its nighttime there.       & Apostrophe   \\ \hline \rowcolor{mygray}
7 & I'd say you could help Zhou Yu. He's either unable to create a non-broken hit or he's cheating, exploring low-wage workers. What do you think?       & Word Confusion   \\ \hline
8 & It just don't work       & SV Agreement   \\ \hline \rowcolor{mygray}
9 & I have a friend from the US. We have a conversation and I don't know the word \textit{bangus} in English. So it was hard for me to communicate with her.       & Verb Form   \\ \hline
\end{tabular}
\caption{\label{tb:ex_table}
Examples user utterances with error type from ErAConD dataset.}
\end{table*}

One of our key goals in developing an error correction model using the proposed dataset is to enable users to focus on specific language skills on which they wish to improve. Since we are dealing with online chat conversations, our data is more casual than the more formal written data seen in previous GEC datasets. Moreover, because our data consists of human-machine conversations involving English language learners of intermediate level, users are assumed to know basic English grammar. Therefore, we wanted to give users the flexibility of choosing to limit feedback, such as only receiving feedback on major lexical and syntactic errors. Specifically, we want to avoid overwhelming users with an excessive number of proposed corrections, and to enable users to improve their conversational skills by first focusing on their most serious errors. Importantly, suggesting an excessive number of corrections could overwhelm a less proficient user or possibly irritate a more proficient participant, resulting in reduced user enjoyment and engagement \citep{koltovskaia2020student}. To that end, we organized our annotated corrections into a 3-level structure based on a perceived ranking of how errors impact the ability of interlocutors to understand what the user is saying, as shown in Table \ref{tb:cat_errors}. As such, we focus primarily on lexical, syntactic and usage errors \citep{ferris2011treatment, touchie1986second}, while leaving mechanical errors to the lowest-priority category. Error priority tags are attached to each edit proposed by the annotators automatically in a post-processing step using a modified version of the ERRANT toolkit \citep{bryant2017automatic}.

For Level 1, our logic is that conversational partners are generally still able to understand a message when it is missing sentence-final punctuation or when a word is not properly capitalized. Because they are of at least intermediate English proficiency, participants can be assumed to know the underlying rules related to punctuation and capitalization; their errors result rather from inattentiveness \citep{sermsook2017analysis} and the informal nature of the conversational genre \citep{cohen1976toward}. Consider Ex.1 in Table \ref{tb:ex_table}: the syntactic structure of the sentence makes it clear that the user is listing names of actors despite the lack of capitalization and punctuation.

For Level 2, our logic is that interlocutors are likely able to understand a message despite usage of acronyms, abbreviations, non-English internet slang, or a missing apostrophe. An example of such non-English internet slang is shown in Ex. 2 in Table \ref{tb:ex_table}. The use of such forms in text-based online conversation is to be expected, since these types of abbreviations are common in all student writing \citep{purcell2013impact, thangaraj2015influence}. However, such cases could potentially lead to misunderstanding, especially when conversing with someone of a different generation or linguistic background. Therefore, we categorize these non-standard forms as moderate \say{errors} (though they are not errors in the traditional sense). We do not consider these non-standard forms as significant because our assumption is that the writer intentionally chose to use these forms for brevity and in the spirit of informality common in online chat \citep{forsythand2007lexical}.


Finally, we include errors which are likely to result misunderstanding or misinterpretation of a message in Level 3 . As we can see in Ex. 7 in Table \ref{tb:ex_table}, the user incorrectly uses the term \textit{non-broken} instead of \textit{unbroken}, and \textit{exploring} instead of \textit{exploiting}. These lexical errors, particularly the latter, are likely to result in misinterpretation of the speaker's intended meaning. Similarly, the user makes a subject-verb agreement error in Ex. 8 and a verb tense error in Ex. 9. In the former, the user mistakenly uses a plural verb for a singular subject, while in the latter, the user uses a present tense verb when a past tense verb is needed. Because these errors relate to some of the most fundamental rules in English grammar, such errors must be addressed promptly. Thus, we treat these errors as \say{significant} in our annotation scheme.

\section{Dataset Statistics}\label{sec:datastat}



\begin{table}[bht]
\small
\centering
\begin{tabular}{lr}
\hline
Dialogs & 186 \\
User turns & 1735 \\
User sentences (source) & 2454 \\
Word tokens (source) & 24616\\
Word types & 2860\\
Error annotations & 2346.5\\
Level 3 error annotations & 684.5 \\
\# of turns per dialog & 9.33 \\
\# of sentences per turn (source) & 1.41 \\
\# of tokens per turn (source) & 14.19 \\
\# of error annotations per turn & 1.35 \\
\# of Level 3 error annotations per turn & 0.39 \\
\# of Level 3 error annotations per 100 tokens & 2.78 \\
\hline
\end{tabular}
\caption{\label{tb:overview} Overview of ErAConD dataset.}
\end{table}

Table \ref{tb:overview} reports statistics related to the composition of the ErAConD  dataset. All statistics are based on user turns; we omit turns generated by our dialog system, as these are not relevant to training a GEC system to provide feedback to users. Additionally, we exclude utterances which include only stop phrases (i.e. \say{stop}, \say{goodbye}, etc.) since these are intended to terminate the conversation. 
Our 3-level structure is reflected in our modified ERRANT \citep{bryant2019bea} toolkit and M2 format. Error type tags are generated from annotated parallel data automatically with our modified version of ERRANT\ifaclfinal\footnote{Code is available at \url{https://github.com/yuanxun-yx/errant}}\else\footnote{Code will be available on GitHub upon acceptance.}\fi, and related figures are averaged across multiple annotators. Inspired by \citet{rozovskaya2021good}, our version of ERRANT also enables users to provide grammatically equivalent edits (i.e. changing ``I'm'' to ``I am''), so that ERRANT can recognize them as identical edits.


As shown in Table \ref{tb:overview}, Level 3 edits account for 29.17\% of all errors, which supports the necessity of our proposed categorization feature. The error distribution in our dataset is comparable to that of essay-based GEC datasets, according to statistics provided in \citet{bryant2019bea}, with the exception of spelling and morphological (inflection) errors, which are substantially higher. While the higher rate of spelling errors is unsurprising in a conversation dataset, the difference in morphological errors warrants further investigation. 

\section{Grammar Error Correction Model}
To demonstrate the utility of our proposed dataset in improving GEC for the open-domain dialog setting, we use the ErAConD dataset to fine-tune a state-of-the-art GEC model, GECToR \citep{omelianchuk2020gector}, which we then test on held-out dialog data. Our results show that ErAConD is useful for adapting GEC models to open-domain dialog. 

\subsection{Training process}
To train a model catered to conversations, we fine-tune the GECToR model\footnote{\url{https://github.com/grammarly/gector##pretrained-models}} proposed by \citet{omelianchuk2020gector} on our collected data. The GECToR model is a grammatical error correction model that generates a set of encoded edit operations to correct the input text rather than directly outputting corrected text. In other words, rather than outputting superficially corrected text, the model outputs the operations necessary to convert the uncorrected text to its corrected version. This set of encoded edit operations can then be applied to the original uncorrected text in a post-processing step to generate the final corrected output \citep{omelianchuk2020gector}. The GECToR model training pipeline starts with a large pre-trained language model (i.e. XLNet or RoBERTa) that is then fine-tuned on both synthetic and collected data. To test our proposed dataset, we further fine-tune this GECToR model on our data.

We only choose to fine-tune the GECToR model using Level 3 edits in our dataset and ignore the Level 1 and 2 edits so that our model can perform better in real-world pedagogical settings. As previously mentioned, overwhelming students with trivial errors, such as punctuation and capitalization, can decrease user enjoyment and engagement \citep{koltovskaia2020student}.
In future work, we plan to train the GECToR model on targeted conversational data across all stages of the pipeline, and determine which errors to present to the user in a post-processing step. We also plan to integrate conversational context.




\subsection{Result and Analysis}\label{sec:expdetail}

\begin{table}[htb]
\scriptsize
\centering
\begin{tabular}{ c|cccccc}
\hline 
\textbf{Setting} & \textbf{TP}& \textbf{FP} & \textbf{FN} & \textbf{Prec} & \textbf{Rec} & \textbf{F${}_{0.5}$} \\ 
\hline
XLNet & 72.4 & 444.6 & 147.2 & 0.140 & 0.330 & 0.158\\
FT XLNet & 27.1 & 13.2 & 191.1 & 0.683 & 0.124 & 0.352\\
\hline
\end{tabular}
\caption{\label{tb:expresult} Performance of GECToR with each setting. Scores are averaged among 5 runs. Table \ref{tb:fiverunresult} provides detailed score of every run. XLNet is the baseline GECToR model based on XLNet, and FT XLNet is the fine-tuned GECToR using level 3 edits.
}
\end{table}

Table \ref{tb:expresult} indicates the efficacy of our data in terms of improving the performance of the GECToR model. The fine-tuned model outperforms the original in terms of F${}_{0.5}$, a metric commonly used in GEC \citep{omelianchuk2020gector}. The significant increase in F${}_{0.5}$ score results from a massive reduction of false positives. In other words, after we fine-tune GECToR on our dataset, the model produces far fewer edits, which helps improve the precision greatly. This is of particular importance in a GEC model, as model precision is considered more important than recall in GEC tasks since false positives could lead to serious confusion in language learners.

Due to the limited size of the dataset, and the uneven distribution of errors in user utterances, we use 5-fold cross-validation to ensure the reliability of our results. We report the average of five cross-validation runs. One note, we modified ERRANT to allow equivalent edits, our reported results on all models might be slightly higher than original ERRANT-based results.

\section{Conclusions and Future Work}

We provide the first  high-quality, fine-grained error-correction conversation dataset  between English second language learner and an educational chatbot. To demonstrate the utility of our dataset, we train and evaluate a SoTA GEC model on the dataset, resulting in a significant improvement in overall model performance for conversational setting. This project lays the groundwork for future work on conversational grammatical error correction (such as adding other dialog domains and incorporating information about the native languages of users) and customized educational dialog system for second language learners.

\section{Ethical Considerations}

Collecting these dialogs for our dataset is difficult in that it requires substantial commitment from participants. In order to provide as large of a dataset as possible, we utilized the services of Amazon Mechanical Turk as previously mentioned. Given ethical concerns in recent years regarding data acquisition through crowdworkers, we verified that the crowdworkers assigned to our tasks were compensated fairly and treated humanely.

The annotators also examined the dataset to ensure that it does not contain personally identifiable information (which was anonymized) or potentially offensive content (which was removed). 


\bibliography{acl2019}

\begin{thebibliography}{32}
\expandafter\ifx\csname natexlab\endcsname\relax\def\natexlab#1{#1}\fi

\bibitem[{Bryant et~al.(2019)Bryant, Felice, Andersen, and
  Briscoe}]{bryant2019bea}
Christopher Bryant, Mariano Felice, {\O}istein~E Andersen, and Ted Briscoe.
  2019.
\newblock {The BEA-2019 shared task on grammatical error correction}.
\newblock In \emph{{Proceedings of the Fourteenth Workshop on Innovative Use of
  NLP for Building Educational Applications}}, pages 52--75.

\bibitem[{Bryant et~al.(2017)Bryant, Felice, and Briscoe}]{bryant2017automatic}
Christopher Bryant, Mariano Felice, and Ted Briscoe. 2017.
\newblock {Automatic Annotation and Evaluation of Error Types for Grammatical
  Error Correction}.
\newblock In \emph{Proceedings of the 55th Annual Meeting of the Association
  for Computational Linguistics (Volume 1: Long Papers)}, pages 793--805.

\bibitem[{Caines et~al.(2020)Caines, Yannakoudakis, Edmondson, Allen,
  P{\'e}rez-Paredes, Byrne, and Buttery}]{caines-etal-2020-teacher}
Andrew Caines, Helen Yannakoudakis, Helena Edmondson, Helen Allen, Pascual
  P{\'e}rez-Paredes, Bill Byrne, and Paula Buttery. 2020.
\newblock \href {https://aclanthology.org/2020.nlp4call-1.2} {{The
  Teacher-Student Chatroom Corpus}}.
\newblock In \emph{Proceedings of the 9th Workshop on NLP for Computer Assisted
  Language Learning}, pages 10--20, Gothenburg, Sweden. LiU Electronic Press.

\bibitem[{Cohen and Robbins(1976)}]{cohen1976toward}
Andrew~D Cohen and Margaret Robbins. 1976.
\newblock Toward assessing interlanguage performance: The relationship between
  selected errors, learners' characteristics, and learners' explanations.
\newblock \emph{Language learning}, 26(1):45--66.

\bibitem[{Davidson et~al.(2019)Davidson, Yu, and Yu}]{davidson2019dependency}
Sam Davidson, Dian Yu, and Zhou Yu. 2019.
\newblock {Dependency Parsing for Spoken Dialog Systems}.
\newblock In \emph{Proceedings of the 2019 Conference on Empirical Methods in
  Natural Language Processing and the 9th International Joint Conference on
  Natural Language Processing (EMNLP-IJCNLP)}, pages 1513--1519.

\bibitem[{Devlin et~al.(2019)Devlin, Chang, Lee, and
  Toutanova}]{devlin2019bert}
Jacob Devlin, Ming-Wei Chang, Kenton Lee, and Kristina Toutanova. 2019.
\newblock {BERT: Pre-training of Deep Bidirectional Transformers for Language
  Understanding}.
\newblock In \emph{Proceedings of the 2019 Conference of the North American
  Chapter of the Association for Computational Linguistics: Human Language
  Technologies, Volume 1 (Long and Short Papers)}, pages 4171--4186.

\bibitem[{Ferris(2011)}]{ferris2011treatment}
Dana Ferris. 2011.
\newblock \emph{Treatment of error in second language student writing}.
\newblock University of Michigan Press.

\bibitem[{Forsythand and Martell(2007)}]{forsythand2007lexical}
Eric~N Forsythand and Craig~H Martell. 2007.
\newblock Lexical and discourse analysis of online chat dialog.
\newblock In \emph{International Conference on Semantic Computing (ICSC 2007)},
  pages 19--26. IEEE.

\bibitem[{Fryer et~al.(2020)Fryer, Coniam, Carpenter, and
  L{\u{a}}pușneanu}]{fryer2020bots}
Luke Fryer, David Coniam, Rollo Carpenter, and Diana L{\u{a}}pușneanu. 2020.
\newblock {Bots for language learning now: Current and future directions}.

\bibitem[{Huang et~al.(2017)Huang, Lee, Kwon, and
  Kim}]{huang2017genietutorplus}
Jin-Xia Huang, Kyung-Soon Lee, Oh-Woog Kwon, and Young-Kil Kim. 2017.
\newblock A chatbot for a dialogue-based second language learning system.
\newblock \emph{CALL in a climate of change: adapting to turbulent global
  conditions--short papers from EUROCALL}, pages 151--156.

\bibitem[{Huang et~al.(2021)Huang, Hew, and Fryer}]{huang2021chatbots}
Weijiao Huang, Khe~Foon Hew, and Luke~K Fryer. 2021.
\newblock {Chatbots for language learning—Are they really useful? A
  systematic review of chatbot-supported language learning}.
\newblock \emph{Journal of Computer Assisted Learning}.

\bibitem[{Koltovskaia(2020)}]{koltovskaia2020student}
Svetlana Koltovskaia. 2020.
\newblock {Student engagement with automated written corrective feedback (AWCF)
  provided by Grammarly: A multiple case study}.
\newblock \emph{Assessing Writing}, 44:100450.

\bibitem[{Li et~al.(2021)Li, Arnold, Yan, Shi, and
  Yu}]{li_arnold_yan_shi_yu_2021}
Yu~Li, Josh Arnold, Feifan Yan, Weiyan Shi, and Zhou Yu. 2021.
\newblock \href {https://doi.org/10.18653/v1/2021.acl-demo.38} {{LEGOEval: An
  Open-Source Toolkit for Dialogue System Evaluation via Crowdsourcing}}.
\newblock \emph{Proceedings of the 59th Annual Meeting of the Association for
  Computational Linguistics and the 11th International Joint Conference on
  Natural Language Processing: System Demonstrations}.

\bibitem[{Liang et~al.(2020)Liang, Chau, Li, Lu, Yu, Zhou, Jain, Davidson,
  Arnold, Nguyen et~al.}]{liang2020gunrock}
Kaihui Liang, Austin Chau, Yu~Li, Xueyuan Lu, Dian Yu, Mingyang Zhou, Ishan
  Jain, Sam Davidson, Josh Arnold, Minh Nguyen, et~al. 2020.
\newblock Gunrock 2.0: A user adaptive social conversational system.
\newblock \emph{arXiv preprint arXiv:2011.08906}.

\bibitem[{Liu et~al.(2019)Liu, Ott, Goyal, Du, Joshi, Chen, Levy, Lewis,
  Zettlemoyer, and Stoyanov}]{liu2019roberta}
Yinhan Liu, Myle Ott, Naman Goyal, Jingfei Du, Mandar Joshi, Danqi Chen, Omer
  Levy, Mike Lewis, Luke Zettlemoyer, and Veselin Stoyanov. 2019.
\newblock Roberta: A robustly optimized bert pretraining approach.
\newblock \emph{arXiv preprint arXiv:1907.11692}.

\bibitem[{N{\'a}plava et~al.(2022)N{\'a}plava, Straka, Strakov{\'a}, and
  Rosen}]{naplava2022czech}
Jakub N{\'a}plava, Milan Straka, Jana Strakov{\'a}, and Alexandr Rosen. 2022.
\newblock {Czech Grammar Error Correction with a Large and Diverse Corpus}.
\newblock \emph{arXiv preprint arXiv:2201.05590}.

\bibitem[{Napoles et~al.(2019)Napoles, N{\u{a}}dejde, and
  Tetreault}]{napoles2019enabling}
Courtney Napoles, Maria N{\u{a}}dejde, and Joel Tetreault. 2019.
\newblock {Enabling robust grammatical error correction in new domains: Data
  sets, metrics, and analyses}.
\newblock \emph{Transactions of the Association for Computational Linguistics},
  7:551--566.

\bibitem[{Napoles et~al.(2017)Napoles, Sakaguchi, and
  Tetreault}]{napoles2017jfleg}
Courtney Napoles, Keisuke Sakaguchi, and Joel Tetreault. 2017.
\newblock {JFLEG: A fluency corpus and benchmark for grammatical error
  correction}.
\newblock \emph{arXiv preprint arXiv:1702.04066}.

\bibitem[{Ng et~al.(2014)Ng, Wu, Briscoe, Hadiwinoto, Susanto, and
  Bryant}]{ng2014conll}
Hwee~Tou Ng, Siew~Mei Wu, Ted Briscoe, Christian Hadiwinoto, Raymond~Hendy
  Susanto, and Christopher Bryant. 2014.
\newblock {The CoNLL-2014 shared task on grammatical error correction}.
\newblock In \emph{Proceedings of the Eighteenth Conference on Computational
  Natural Language Learning: Shared Task}, pages 1--14.

\bibitem[{Omelianchuk et~al.(2020)Omelianchuk, Atrasevych, Chernodub, and
  Skurzhanskyi}]{omelianchuk2020gector}
Kostiantyn Omelianchuk, Vitaliy Atrasevych, Artem Chernodub, and Oleksandr
  Skurzhanskyi. 2020.
\newblock {GECToR--Grammatical Error Correction: Tag, Not Rewrite}.
\newblock \emph{arXiv preprint arXiv:2005.12592}.

\bibitem[{Purcell et~al.(2013)Purcell, Buchanan, and
  Friedrich}]{purcell2013impact}
Kristen Purcell, Judy Buchanan, and Linda Friedrich. 2013.
\newblock The impact of digital tools on student writing and how writing is
  taught in schools.
\newblock \emph{Washington, DC: Pew Research Center}.

\bibitem[{Roller et~al.(2020)Roller, Dinan, Goyal, Ju, Williamson, Liu, Xu,
  Ott, Shuster, Smith, Boureau, and Weston}]{roller2020recipes}
Stephen Roller, Emily Dinan, Naman Goyal, Da~Ju, Mary Williamson, Yinhan Liu,
  Jing Xu, Myle Ott, Kurt Shuster, Eric~M. Smith, Y-Lan Boureau, and Jason
  Weston. 2020.
\newblock Recipes for building an open-domain chatbot.

\bibitem[{Rothe et~al.(2021)Rothe, Mallinson, Malmi, Krause, and
  Severyn}]{rothe2021simple}
Sascha Rothe, Jonathan Mallinson, Eric Malmi, Sebastian Krause, and Aliaksei
  Severyn. 2021.
\newblock {A Simple Recipe for Multilingual Grammatical Error Correction}.
\newblock \emph{arXiv preprint arXiv:2106.03830}.

\bibitem[{Rozovskaya and Roth(2010)}]{rozovskaya2010annotating}
Alla Rozovskaya and Dan Roth. 2010.
\newblock Annotating {ESL} errors: {Challenges and rewards}.
\newblock In \emph{Proceedings of the NAACL HLT 2010 fifth workshop on
  innovative use of NLP for building educational applications}, pages 28--36.

\bibitem[{Rozovskaya and Roth(2019)}]{rozovskaya2019grammar}
Alla Rozovskaya and Dan Roth. 2019.
\newblock {Grammar Error Correction in Morphologically Rich Languages: The Case
  of Russian}.
\newblock \emph{Transactions of the Association for Computational Linguistics},
  7:1--17.

\bibitem[{Rozovskaya and Roth(2021)}]{rozovskaya2021good}
Alla Rozovskaya and Dan Roth. 2021.
\newblock {How Good (really) are Grammatical Error Correction Systems?}
\newblock In \emph{Proceedings of the 16th Conference of the European Chapter
  of the Association for Computational Linguistics: Main Volume}, pages
  2686--2698.

\bibitem[{Sermsook et~al.(2017)Sermsook, Liamnimit, and
  Pochakorn}]{sermsook2017analysis}
Kanyakorn Sermsook, Jiraporn Liamnimit, and Rattaneekorn Pochakorn. 2017.
\newblock {An Analysis of Errors in Written English Sentences: A Case Study of
  Thai EFL Students.}
\newblock \emph{English Language Teaching}, 10(3):101--110.

\bibitem[{Stahlberg and Kumar(2021)}]{stahlberg2021synthetic}
Felix Stahlberg and Shankar Kumar. 2021.
\newblock {Synthetic Data Generation for Grammatical Error Correction with
  Tagged Corruption Models}.
\newblock In \emph{Proceedings of the 16th Workshop on Innovative Use of NLP
  for Building Educational Applications}, pages 37--47.

\bibitem[{Thangaraj and Maniam(2015)}]{thangaraj2015influence}
Shalini~Raj Thangaraj and Mahendran Maniam. 2015.
\newblock {The Influence of Netspeak on students' writing}.
\newblock \emph{Journal of Education and Learning}, 9(1):45--52.

\bibitem[{Touchie(1986)}]{touchie1986second}
Hanna~Y Touchie. 1986.
\newblock {Second language learning errors: Their types, causes, and
  treatment}.
\newblock \emph{JALT journal}, 8(1):75--80.

\bibitem[{Trinh and Rozovskaya(2021)}]{trinh2021new}
Viet~Anh Trinh and Alla Rozovskaya. 2021.
\newblock {New Dataset and Strong Baselines for the Grammatical Error
  Correction of Russian}.
\newblock In \emph{Findings of the Association for Computational Linguistics:
  ACL-IJCNLP 2021}, pages 4103--4111.

\bibitem[{Yang et~al.(2019)Yang, Dai, Yang, Carbonell, Salakhutdinov, and
  Le}]{yang2019xlnet}
Zhilin Yang, Zihang Dai, Yiming Yang, Jaime Carbonell, Russ~R Salakhutdinov,
  and Quoc~V Le. 2019.
\newblock {XLNet: Generalized autoregressive pretraining for language
  understanding}.
\newblock \emph{Advances in neural information processing systems}, 32.

\end{thebibliography}
\bibliographystyle{acl_natbib}

\appendix

\section{Appendices}
\label{sec:appendix}

\subsection{Annotation Exceptions}
Even though they violate the rules of standard English, we left the following types of errors unchanged in our annotated dataset:
\begin{enumerate}
    \item \textit{Utterances that are not complete sentences.} For example, response utterances such as \textit{Yes}, \textit{Very good}, and \textit{Me too} are considered correct in our annotation due to their prevalence in informal dialog, although they are not correct in formal writing.
    \item \textit{Use of common English internet slang and shorthand expressions.} Slang and shorthand expressions such as \textit{lol} (``laugh out loud") and \textit{u} (short for ``you") are not only distinctive to online chat conversations, but also reflective of their casual nature. Additionally, they may be language, culture, and even sub-culture specific. While these terms may not be suitable to a more formal register, they are generally acceptable in the context of informal dialog \citep{forsythand2007lexical}; thus, we do not classify such usage as errors.
\end{enumerate}

\subsection{Dataset Statistics}

\begin{table}[htb]
\small
\centering
\begin{tabular}{ clrr}
\hline 
\textbf{Level} & \textbf{Type}& \textbf{Number} & \textbf{\%} \\ 
\hline
\multirow{ 3 }*{ 1 } 
& PUNCT & 824.5 & 63.28 \\
& ORTH & 478.5 & 36.72 \\
& \textbf{Total} & \textbf{ 1303.0 }&\textbf{ 55.45 }\\
\hline
\multirow{ 9 }*{ 2 } 
& SPELL & 0.5 & 0.14 \\
& PUNCT & 229.5 & 63.31 \\
& PREP & 1.0 & 0.28 \\
& OTHER & 124.5 & 34.34 \\
& NOUN:POSS & 3.5 & 0.97 \\
& NOUN & 2.0 & 0.55 \\
& DET & 0.5 & 0.14 \\
& ADJ & 1.0 & 0.28 \\
& \textbf{Total} & \textbf{ 362.5 }&\textbf{ 15.43 }\\
\hline
\multirow{ 24 }*{ 3 } 
& WO & 9.5 & 1.39 \\
& VERB:TENSE & 37.5 & 5.48 \\
& VERB:SVA & 19.0 & 2.78 \\
& VERB:INFL & 1.0 & 0.15 \\
& VERB:FORM & 37.5 & 5.48 \\
& VERB & 40.0 & 5.84 \\
& SPELL & 115.5 & 16.87 \\
& SPACE & 11.0 & 1.61 \\
& PRON & 34.0 & 4.97 \\
& PREP & 69.0 & 10.08 \\
& PART & 4.0 & 0.58 \\
& OTHER & 110.0 & 16.07 \\
& NOUN:POSS & 3.5 & 0.51 \\
& NOUN:NUM & 35.5 & 5.19 \\
& NOUN:INFL & 2.5 & 0.37 \\
& NOUN & 35.5 & 5.19 \\
& MORPH & 28.0 & 4.09 \\
& DET & 57.0 & 8.33 \\
& CONTR & 4.0 & 0.58 \\
& CONJ & 3.5 & 0.51 \\
& ADV & 15.0 & 2.19 \\
& ADJ:FORM & 2.5 & 0.37 \\
& ADJ & 9.5 & 1.39 \\
& \textbf{Total} & \textbf{ 684.5 }&\textbf{ 29.13 }\\
\hline
\end{tabular}
\caption{\label{tb:typedistfull}
Error type distribution. 
}
\end{table}

As described in Section \ref{sec:datastat}, Table \ref{tb:typedistfull} shows the type distribution of edit type in ErAConD. Type labels were generated using our version of ERRANT. Levels of edits were first generated by ERRANT, and then manually checked to label Type 2 edits that are hard to be recognized by code (non-English Internet slangs, acronyms and abbreviations). 
To take all annotators into consideration, the number of edits was averaged among multiple annotators. 

The statistics give us several important insights. 
First, the number of \say{significant} errors is slightly higher than in written GEC datasets, such as NUCLE. This result shows that grammatical errors are relatively rare in both the conversational and written domain.
Additionally, the average length of each sentence is significantly shorter than written GEC datasets. Finally, the error rate data supports our tiered categorization of errors, as the frequency of errors would be much higher than non-conversational datasets if all less significant errors, such as capitalization and punctuation, were included. 

\subsection{Experimental Results}

Table \ref{tb:fiverunresult} is the full version of Table \ref{tb:expresult}. Some details of experiment are mentioned at Section \ref{sec:expdetail}. 20\% of the dialogs were chosen randomly for the test set and the rest were used for training. Then 5-fold cross-validation was applied and the whole process was run 5 times in total, so as to observe the reliability of our results. We used the recommended parameters of XLNet to train and test GECToR.
From the table we can see that the variance of performance among these runs is small.
    The distribution of Level 3 edits in test and train sets for each run is also represented in Table \ref{tb:typedistfive}. 

\begin{table*}[tb!]
\small
\centering
\begin{tabular}{ccccccccc}
\hline
\textbf{Run No.} &
\textbf{Setting} & 
\textbf{TP}& \textbf{FP} & \textbf{FN} & \textbf{Prec} & \textbf{Rec} & \textbf{F${}_{0.5}$} \\ 
\hline\multirow{ 2 }*{ 1 }
& XLNet & 54 & 395 & 157 & 0.120 & 0.256 & 0.135\\
& FT XLNet & 21.4 & 6.8 & 184.6 & 0.759 & 0.104 & 0.336\\
\hline\multirow{ 2 }*{ 2 }
& XLNet & 71 & 506 & 134 & 0.123 & 0.346 & 0.141\\
& FT XLNet & 24.4 & 11.0 & 179.6 & 0.690 & 0.120 & 0.353\\
\hline\multirow{ 2 }*{ 3 }
& XLNet & 77 & 437 & 168 & 0.150 & 0.314 & 0.167\\
& FT XLNet & 25.4 & 14.6 & 219.6 & 0.637 & 0.104 & 0.313\\
\hline\multirow{ 2 }*{ 4 }
& XLNet & 74 & 404 & 146 & 0.155 & 0.336 & 0.173\\
& FT XLNet & 22.6 & 10.4 & 196.4 & 0.686 & 0.103 & 0.321\\
\hline\multirow{ 2 }*{ 5 }
& XLNet & 86 & 481 & 131 & 0.152 & 0.396 & 0.173\\
& FT XLNet & 41.6 & 23.2 & 175.4 & 0.642 & 0.192 & 0.437\\
\hline\multirow{ 2 }*{ Avg. }
& XLNet & 72.4 & 444.6 & 147.2 & 0.140 & 0.330 & 0.158\\
& FT XLNet & 27.1 & 13.2 & 191.1 & 0.683 & 0.124 & 0.352\\
\hline
\end{tabular}
\caption{\label{tb:fiverunresult}
Performance of GECToR with each setting in 5 runs. 
}
\end{table*}

\begin{table*}[!b]
\small
\centering
\begin{tabular}{ lrrrrrrrrrr}
\hline 
\multirow{2}*{\textbf{Type}} &
\multicolumn{2}{c}{\textbf{1}} &
\multicolumn{2}{c}{\textbf{2}} &
\multicolumn{2}{c}{\textbf{3}} &
\multicolumn{2}{c}{\textbf{4}} &
\multicolumn{2}{c}{\textbf{5}} 
\\ 
&
\textbf{Test} & \textbf{Train} &
\textbf{Test} & \textbf{Train} &
\textbf{Test} & \textbf{Train} &
\textbf{Test} & \textbf{Train} &
\textbf{Test} & \textbf{Train} 
\\ \hline
WO & 1.03 & 1.48 & 1.23 & 1.42 & 1.14 & 1.45 & 1.74 & 1.32 & 0.87 & 1.49 \\
VERB:TENSE & 8.28 & 4.73 & 5.33 & 5.51 & 5.68 & 5.43 & 2.17 & 6.15 & 6.11 & 5.35 \\
VERB:SVA & 4.14 & 2.41 & 2.46 & 2.84 & 2.27 & 2.90 & 0.43 & 3.25 & 3.06 & 2.72 \\
VERB:INFL & 0.34 & 0.09 & 0.00 & 0.18 & 0.38 & 0.09 & 0.00 & 0.18 & 0.00 & 0.18 \\
VERB:FORM & 4.83 & 5.65 & 5.33 & 5.51 & 4.55 & 5.70 & 5.22 & 5.53 & 6.55 & 5.26 \\
VERB & 6.21 & 5.75 & 4.92 & 6.04 & 6.06 & 5.79 & 6.09 & 5.79 & 4.37 & 6.14 \\
SPELL & 15.17 & 17.33 & 17.21 & 16.80 & 17.42 & 16.74 & 19.13 & 16.42 & 18.78 & 16.49 \\
SPACE & 1.38 & 1.67 & 1.64 & 1.60 & 3.03 & 1.27 & 1.30 & 1.67 & 0.87 & 1.75 \\
PRON & 8.97 & 3.89 & 4.51 & 5.07 & 4.92 & 4.98 & 3.04 & 5.36 & 3.93 & 5.18 \\
PREP & 9.31 & 10.29 & 11.07 & 9.87 & 10.23 & 10.05 & 8.70 & 10.36 & 13.97 & 9.30 \\
PART & 1.38 & 0.37 & 0.82 & 0.53 & 0.38 & 0.63 & 0.00 & 0.70 & 1.31 & 0.44 \\
OTHER & 16.90 & 15.85 & 16.39 & 16.00 & 15.53 & 16.20 & 17.83 & 15.72 & 10.48 & 17.19 \\
NOUN:POSS & 0.00 & 0.65 & 0.41 & 0.53 & 0.38 & 0.54 & 0.87 & 0.44 & 0.44 & 0.53 \\
NOUN:NUM & 5.52 & 5.10 & 8.61 & 4.44 & 5.30 & 5.16 & 7.39 & 4.74 & 6.55 & 4.91 \\
NOUN:INFL & 0.34 & 0.37 & 0.41 & 0.36 & 1.52 & 0.09 & 0.87 & 0.26 & 0.87 & 0.26 \\
NOUN & 1.38 & 6.21 & 5.74 & 5.07 & 3.41 & 5.61 & 4.78 & 5.27 & 2.62 & 5.70 \\
MORPH & 2.41 & 4.54 & 2.05 & 4.53 & 3.79 & 4.16 & 5.65 & 3.78 & 3.49 & 4.21 \\
DET & 7.59 & 8.53 & 7.38 & 8.53 & 9.47 & 8.05 & 7.83 & 8.43 & 11.79 & 7.63 \\
CONTR & 0.00 & 0.74 & 0.41 & 0.62 & 0.38 & 0.63 & 1.30 & 0.44 & 0.87 & 0.53 \\
CONJ & 0.34 & 0.56 & 0.41 & 0.53 & 0.00 & 0.63 & 0.87 & 0.44 & 0.00 & 0.61 \\
ADV & 2.07 & 2.22 & 1.64 & 2.31 & 2.65 & 2.08 & 2.61 & 2.11 & 1.31 & 2.37 \\
ADJ:FORM & 0.34 & 0.37 & 0.82 & 0.27 & 0.38 & 0.36 & 0.43 & 0.35 & 0.87 & 0.26 \\
ADJ & 2.07 & 1.20 & 1.23 & 1.42 & 1.14 & 1.45 & 1.74 & 1.32 & 0.87 & 1.49 \\
\hline
\end{tabular}
\caption{\label{tb:typedistfive}
Level 3 error type distribution (\%) in train and test sets of 5 runs.
}
\end{table*}

\end{document}